\documentclass{article}
\usepackage[utf8]{inputenc}
\usepackage{xcolor,graphicx}
\usepackage{booktabs}
\usepackage{authblk}
\usepackage{setspace}
\usepackage{amssymb, amsfonts, amsmath}
\usepackage{algorithm}
\usepackage{algpseudocode}

\usepackage[margin=1.25in]{geometry}
\graphicspath{ {./figures/} }
\usepackage{subcaption}
\usepackage{amsmath}
\usepackage{lineno}

\title{Optimal hyperspectral undersampling strategy\\ for satellite imaging}

\author[1,2*$\dag$]{Vita V. Vlasova}
\author[4,7$\dag$]{Vladimir G. Kuzmin}
\author[4,6$\dag$]{Maria S. Varetsa}
\author[3,5]{Natalia A. Ibragimova}
\author[2,3,4]{Oleg Y. Rogov}
\author[1]{Elena V. Lyapuntsova}

\affil[1]{Bauman Moscow State Technical University, Moscow, Russia.}
\affil[2]{Artificial Intelligence Research Institute, Moscow, Russia.}
\affil[3]{Moscow Technical University of Communications and Informatics, Moscow, Russia.}
\affil[4]{VeinCV LLC, Moscow, Russia}
\affil[5]{2GIS, Moscow, Russia}
\affil[6]{Nanosemantics LLC, Moscow, Russia}
\affil[7]{Skoltech, Moscow, Russia}
\affil[*]{Address correspondence to: \texttt{vlasova@airi.net}, \texttt{lev86@bmstu.ru}}
\affil[$\dag$]{Equal contribution.}

\date{}

\begin{document}

\maketitle

\begin{abstract}
Hyperspectral image (HSI) classification presents significant challenges due to the high dimensionality, spectral redundancy, and limited labeled data typically available in real-world applications. To address these issues and optimize classification performance, we propose a novel band selection strategy known as Iterative Wavelet-based Gradient Sampling (IWGS). This method incrementally selects the most informative spectral bands by analyzing gradients within the wavelet-transformed domain, enabling efficient and targeted dimensionality reduction. Unlike traditional selection methods, IWGS leverages the multi-resolution properties of wavelets to better capture subtle spectral variations relevant for classification. The iterative nature of the approach ensures that redundant or noisy bands are systematically excluded while maximizing the retention of discriminative features. We conduct comprehensive experiments on two widely-used benchmark HSI datasets: Houston 2013 and Indian Pines. Results demonstrate that IWGS consistently outperforms state-of-the-art band selection and classification techniques in terms of both accuracy and computational efficiency. These improvements make our method especially suitable for deployment in edge devices or other resource-constrained environments, where memory and processing power are limited. In particular, IWGS achieved an overall accuracy up to  97.8\% on Indian Pines for selected classes, confirming its effectiveness and generalizability across different HSI scenarios.
\end{abstract}

\section{Introduction}\label{Introduction}

Hyperspectral imaging (HSI) has revolutionized modern computationally abundant domains such as remote sensing~\cite{rs13112181,Belov2023,9078584}, video processing~\cite{BrednevHS}, agriculture~\cite{Zolotukhina2024} through detailed analysis of land cover and environmental conditions~\cite{Guzeva_2022}. Unlike conventional imaging systems, HSI acquires hundreds of adjacent, narrow spectral bands spanning the electromagnetic spectrum~\cite{wang2023hyperspectral, ran2016bands}. This fine spectral resolution allows for the precise identification and classification of surface materials by capturing subtle spectral signatures not visible in RGB imagery~\cite{ikeuchi2021computer, mateen2018role}. As a result, HSI has found widespread utility in diverse applications such as precision agriculture, mineral exploration, urban mapping, and ecological monitoring~\cite{barbedo2023review, tao2017automatic, vadrevu2022remote}.

Nevertheless, the inherently high dimensionality of hyperspectral data introduces substantial computational challenges in classification tasks. The large number of spectral channels increases processing time and complexity, necessitating methods that effectively reduce dimensionality while retaining key spectral and spatial information~\cite{moharram2023land}. A widely adopted approach to address this issue is band selection~\cite{cai2019bs,wang2019attend,yang2024lidar}, which involves identifying and retaining only the most informative spectral bands. By focusing on the most discriminative wavelengths, band selection reduces redundancy and improves efficiency, which is particularly important for real-time systems and edge devices with limited resources~\cite{sun2019hyperspectral}. Concurrently, the emergence of attention-based deep learning architectures, such as Vision Transformers (ViTs)~\cite{vaswani2017attention, aleissaee2023transformers}, has significantly advanced the field of image understanding. ViTs excel in modeling long-range dependencies by encoding global context across image patches, and have outperformed convolutional neural networks (CNNs) in various computer vision domains~\cite{hong2021spectralformer, he2019hsi}. However, the extensive computational and memory demands of ViTs limit their practicality in remote sensing scenarios, especially in environments with constrained hardware resources.

Recent progress in State Space Models (SSMs) presents a promising alternative to transformers. These models support parallelized sequence processing and can efficiently model long-range dependencies with lower complexity. The Mamba architecture exemplifies this trend, delivering competitive visual understanding capabilities with linear scalability and improved efficiency~\cite{gu2023mamba, zhu2024vision}. Inspired by this balance between performance and computational thrift, we propose to combine the strengths of CNNs and iterative undersampling~\cite{RAZUMOV202337} to address the unique requirements of hyperspectral data. We introduce a hybrid neural architecture designed to boost HSI classification performance through a synergistic fusion of spatial and spectral learning. Our model employs a bidirectional network structure to process hyperspectral inputs effectively, combining CNN-driven spatial feature extraction with lightweight attention-inspired spectral modeling. Drawing from Mamba's efficient design principles, the the non-linear SSM delivers accurate classification with reduced memory and computational overhead. This enables thorough analysis of both local and global structures within HSI data, without incurring the substantial cost of full transformer models. We validate the proposed method with the Indian Pines dataset demonstrating its superiority over state-of-the-art models in both classification accuracy and computational performance. The model significantly lowers GPU memory requirements, CPU utilization, and inference latency, positioning it as an effective solution for real-world applications that demand both precision and efficiency.

Our work makes the following threefold contributions:
\begin{itemize} 
    \item We propose IGWS (Iterative Wavelet-based Gradient Sampling, a novel undersampling strategy that efficiently balances class distributions by preserving informative minority-class samples.
    \item Our model achieves notable computational advantages over conventional architectures, including RNN-, CNN-, and ViT-based models, by minimizing memory footprint and processing demands, which is crucial for scalable remote sensing workflows.
    \item We empirically validate the model using public benchmark consistently surpassing leading transformer-based baselines in classification performance including adversarial robustness.
\end{itemize}

\noindent The remainder of the paper is organized as follows: Section~\ref{sec:Related Work} surveys related literature on HSI classification. Section~\ref{sec:methodology} outlines our proposed methodology, including the algorithm and the band selection strategy. Section~\ref{sec:experiments} presents the experimental setup, results, and analysis. Finally, Section~\ref{sec:conclusion} concludes the paper and discusses future research directions.

\section{Related Work}\label{sec:Related Work}

This section outlines key developments in hyperspectral image (HSI) classification, with a focus on recent advances in deep learning techniques. We highlight the contributions of convolutional neural networks (CNNs), transformer-based architectures, and the emerging class of SSMs, each offering unique advantages in handling the complexities of HSI data.

\subsection{Deep Learning in Hyperspectral Image Analysis}\label{sec:DLCNN}

Deep learning has significantly advanced the field of HSI analysis and classification by enabling more effective extraction and interpretation of spatial and spectral features. Architectures such as CNNs, recurrent neural networks (RNNs), and generative adversarial networks (GANs) have been adapted to suit the high-dimensional and information-rich nature of hyperspectral data~\cite{zhou2019learning, yu2017convolutional, mou2017deep, zhu2018generative, paoletti2018capsule}.

\textbf{CNN-Based Models.} CNNs have proven particularly effective due to their ability to learn spatial hierarchies and capture localized patterns critical for HSI tasks, specifically for segmentation~\cite{Podoprigorova2024} and object detection~\cite{Kanev2024,Kanev2023}. Initial models like the 2D CNN~\cite{yang2018hyperspectral} utilized conventional convolutional layers to process spatial dimensions, integrating pooling and normalization to manage data complexity. The introduction of R-2D-CNN further refined this approach with residual connections, enabling deeper architectures while preserving spatial resolution.

Building upon 2D frameworks, the 3D CNN~\cite{yang2018hyperspectral} extended the receptive field to the spectral domain, using volumetric convolutions to simultaneously extract spectral-spatial features. Although this design improved feature richness, it also introduced higher computational costs due to the complexity of 3D operations.

\textbf{Advanced CNN Models.} To address the scalability challenges of early CNNs, models such as M3D-DCNN~\cite{he2017multi} introduced a multi-scale, end-to-end design that jointly processes 2D spatial and 1-D spectral features. By accommodating multiple feature resolutions, M3D-DCNN effectively balances detail preservation and computational efficiency, making it suitable for large-scale HSI applications.

Despite their success in modeling spatial structure, standalone CNNs may fail to fully capture long-range spectral dependencies critical to hyperspectral analysis. As a result, hybrid models that integrate CNNs with sequential or attention-based mechanisms have been proposed to bridge this gap.

\textbf{RNN-Based Models.} Recurrent models offer a natural fit for sequential spectral data. Mou et al.~\cite{mou2017deep} introduced an architecture incorporating the parametric rectified tanh (PRetanh) activation and customized gated recurrent units to capture spectral continuity. These models demonstrated effective sequence modeling, but their inherently sequential processing can lead to inefficiencies in large-scale or real-time settings.

\subsection{Transformers in Hyperspectral Image Classification}\label{sec: Transformer}

Transformer-based models have redefined the landscape of HSI classification by utilizing self-attention to model both local and long-range dependencies. Originally designed for natural language processing, Vision Transformers (ViTs)~\cite{vaswani2017attention} have been successfully adapted for visual tasks, offering new capabilities in spectral-spatial understanding.

\textbf{Transformer Architectures.} Models such as SpectralFormer~\cite{hong2021spectralformer} leverage cross-layer skip connections to enhance spectral feature representation, eliminating the need for complex preprocessing. HSI-BERT~\cite{he2019hsi} introduced a bidirectional transformer framework tailored to hyperspectral data, improving classification accuracy through the joint modeling of spectral and spatial features.

\textbf{Transformer Modifications.} More recent architectures—including Deep ViT~\cite{zhou2021deepvit}, T2T~\cite{yuan2021tokens}, LeViT~\cite{graham2021levit}, and HiT~\cite{yang2022hyperspectral}—further expand transformer capacity for HSI by enhancing both spectral and spatial modeling. These models excel in capturing global context but face limitations in deployment due to their substantial memory and processing requirements.

\textbf{Hybrid CNN-Transformer Models.} The fusion of CNNs with transformer modules has led to novel hybrid architectures. For instance, the HiT model~\cite{yang2022hyperspectral} integrates CNN-based convolutional layers within a transformer pipeline, combining local feature extraction with global attention. Similarly, the multiscale convolutional transformer~\cite{jia2022multiscale} and the spectral-spatial feature tokenization transformer (SSFTT)~\cite{sun2022spectral} demonstrate that combining CNNs with self-attention can significantly enhance spectral-spatial representation in HSI classification. Although these models set new benchmarks in classification accuracy, their computational cost remains a barrier to practical deployment, particularly in edge-computing or remote sensing scenarios with limited resources.

\textbf{State Space Models} have recently emerged as a scalable alternative to transformers in vision applications. The Mamba model~\cite{gu2023mamba} exemplifies this trend by providing efficient long-range dependency modeling with linear computational complexity, positioning itself as a high-performance yet resource-efficient option for vision tasks.

Mamba eliminates the need for resource-intensive attention mechanisms, instead relying on a compact recurrent formulation that scales linearly with sequence length. This advantage makes it particularly attractive for applications involving high-resolution imagery and dense data, such as HSI~\cite{fu2022hungry}.

Integrating Mamba-style SSMs with CNN backbones enables models to retain spatial structure while achieving spectral modeling at significantly lower computational costs. This synergy enhances the potential for deploying deep learning models in constrained environments without sacrificing performance.

\subsection{Synthesis and Outlook}

The collective evolution of CNNs, transformers, and SSMs reflects the dynamic landscape of HSI classification. CNNs, especially 3D CNNs and their multi-scale variants, have laid the groundwork for robust spatial-spectral modeling. Transformers have introduced global context and spectral dependency modeling at unprecedented levels, while hybrid approaches leverage the best of both worlds. Most recently, Mamba-based models offer a compelling path forward by optimizing memory and compute efficiency without compromising accuracy.

As HSI research advances, the convergence of these approaches is expected to yield increasingly powerful models that combine computational efficiency with classification precision, making hyperspectral analytics more accessible for real-world applications.

\section{Methodology}
\label{sec:methodology}
Our proposed framework addresses hyperspectral image (HSI) classification by integrating a learnable band selection mechanism with an efficient hybrid neural architecture. Central to this framework is the Iterative Wavelet-based Gradient Sampling (IWGS) algorithm, which selects informative spectral bands in a task-specific manner.

\subsection{Iterative Wavelet-based Gradient Sampling (IWGS)}

Let $\mathbf{X} \in \mathbb{R}^{H \times W \times B}$ denote the input hyperspectral cube, where $H$ and $W$ are the spatial dimensions and $B$ is the total number of spectral bands. The objective of IWGS is to select a subset of $N_s$ spectral bands that minimizes the classification loss using a model $\mathcal{S}(\cdot)$.

We introduce a binary selection vector $\mathbf{w} \in \{0,1\}^B$, where $\mathbf{w}_j = 1$ indicates that the $j$-th band is selected. Optimization progresses iteratively by updating $\mathbf{w}$ to select the band that produces the greatest reduction in classification loss.

The wavelet transform is denoted by an operator as $\mathcal{W}(\cdot)$ and its inverse as $\mathcal{W}^{-1}(\cdot)$. The loss function is therefore $\mathcal{L}_{\text{target}}(Y, \hat{Y})$, where $Y$ are the ground truth labels and $\hat{Y}$ are the predictions.

\begin{algorithm}[t]
\caption{Iterative Wavelet-based Gradient Sampling (IWGS)}
\label{alg:iwgs}
\begin{algorithmic}[1]
\Require Hyperspectral cube $\mathbf{X}$, labels $Y$, classifier $\mathcal{S}(\cdot)$, wavelet transform $\mathcal{W}$, inverse wavelet transform $\mathcal{W}^{-1}$, number of selected bands $N_s$
\Ensure Band selection vector $\mathbf{w} \in \{0,1\}^B$
\State Initialize $\mathbf{w} \gets \mathbf{0}$; select central band: $\mathbf{w}_{B/2} \gets 1$
\For{$n = 1$ to $N_s$}
    \State Compute wavelet domain representation: $\mathbf{X}_W \gets \mathcal{W}(\mathbf{X})$
    \State Reconstruct masked input: $\hat{\mathbf{X}}_{\mathbf{w}} \gets \mathcal{W}^{-1}(\mathbf{X}_W \cdot \mathbf{w})$
    \State Predict: $\hat{Y} \gets \mathcal{S}(\hat{\mathbf{X}}_{\mathbf{w}})$
    \State Compute loss: $\mathcal{L} \gets \mathcal{L}_{\text{target}}(Y, \hat{Y})$
    \State Compute gradient: $\nabla_{\mathbf{w}} \mathcal{L}$
    \State Identify next band: $j^\star \gets \arg\min_{j : \mathbf{w}_j = 0} \left| \frac{\partial \mathcal{L}}{\partial \mathbf{w}_j} \right|$
    \State Update selection vector: $\mathbf{w}_{j^\star} \gets 1$
\EndFor
\end{algorithmic}
\end{algorithm}

\subsection{Adversarial Perturbations}
Adversarial perturbations in HSP are crafted to exploit the vulnerability of deep models to minor, input-specific changes that cause misclassification in a variety of applications~\cite{Korzh2025,8725896}. PGD is a first-order iterative attack method that perturbs input data within a defined $\ell_p$-norm constraint, guiding the model's prediction away from the ground truth label. In the context of hyperspectral data, the high-dimensional spectral signature of each pixel is sensitive to both noise and adversarial interference. The Iterative Gradient-based Wavelet Sampling (IGWS) method is evaluated under these conditions, specifically focusing on its robustness when exposed to compound perturbations involving atmospheric noise and limited training data. This setup reflects practical deployment scenarios, where data collection is often constrained and environmental interference is inevitable. The resilience of IGWS is validated by tracking changes in the Kappa statistic across patch scales, revealing degradation patterns and identifying robustness trends.
\subsection{Noise Model}

Let $\mathbf{X} \in \mathbb{R}^{H \times W \times B}$ denote a hyperspectral image, where $H$ and $W$ are the spatial dimensions, and $B$ is the number of spectral bands. Each pixel $\mathbf{x}_{i} \in \mathbb{R}^{B}$ corresponds to a spectral signature at spatial location $i$. The objective is to learn a function $f_\theta: \mathbb{R}^B \rightarrow \mathcal{Y}$, parameterized by $\theta$, that maps input spectra to a set of land cover class labels $\mathcal{Y} = \{1, \dots, C\}$.

\vspace{0.5em}
\paragraph{Classification Objective.}
Given a hyperspectral image cube $\mathbf{X} \in \mathbb{R}^{H \times W \times B}$ and its corresponding ground-truth label matrix $Y \in \mathcal{Y}^{H \times W}$, the classification problem is defined as learning a mapping $\mathcal{S}: \mathbb{R}^{H \times W \times B} \rightarrow \mathcal{Y}^{H \times W}$ that minimizes the expected loss over the data distribution:
\begin{equation}
\min_{\theta} \; \mathbb{E}_{(\mathbf{X}, Y) \sim \mathcal{D}} \; \mathcal{L}_{\text{target}} \big(Y, \mathcal{S}_\theta(\mathbf{X})\big),
\end{equation}
where $\mathcal{S}_\theta$ denotes a spectral-spatial classifier parameterized by $\theta$, and $\mathcal{L}_{\text{target}}$ is a task-specific loss function (e.g., pixel-wise cross-entropy). The goal is to predict the label at each spatial location using the full spectral information or a selected subset of informative bands, as determined by sampling strategies like IWGS.

\paragraph{Adversarial PGD Perturbations.}
Beyond stochastic noise, we also consider adversarial perturbations crafted to maximally degrade classifier performance. Using the Projected Gradient Descent (PGD) attack~\cite{madry2019deeplearningmodelsresistant}, the adversarial sample is generated as:
\[
\mathbf{x}_i^{\text{adv}} = \Pi_{\mathcal{B}_\epsilon(\mathbf{x}_i)} \left( \mathbf{x}_i + \alpha \cdot \text{sign}(\nabla_{\mathbf{x}} \mathcal{L}(f_\theta(\mathbf{x}), y_i)) \right),
\]
where $\Pi_{\mathcal{B}_\epsilon(\mathbf{x}_i)}$ denotes projection onto the $\ell_\infty$-ball of radius $\epsilon$ centered at $\mathbf{x}_i$, and $\alpha$ is the step size. The perturbation is bounded but optimized to fool the classifier, simulating worst-case spectral distortion scenarios in deployment environments.

\paragraph{Robust Classification Objective.}
To account for both atmospheric noise and adversarial perturbations, we define a robust objective over the hyperspectral cube $\mathbf{X}$ and label matrix $Y \in \mathcal{Y}^{H \times W}$. Let $\boldsymbol{\eta} \in \mathbb{R}^{H \times W \times B}$ denote additive atmospheric noise and $\boldsymbol{\delta} \in \mathbb{R}^{H \times W \times B}$ an adversarial perturbation constrained within an $\ell_\infty$-ball. The robust classification problem is formulated as:
\begin{equation}
\min_{\theta} \; \mathbb{E}_{(\mathbf{X}, Y) \sim \mathcal{D}} \left[ \max_{\|\boldsymbol{\delta}\|_\infty \leq \epsilon} \; \mathcal{L}_{\text{target}}\left( Y, \mathcal{S}_\theta(\mathbf{X} + \boldsymbol{\eta} + \boldsymbol{\delta}) \right) \right],
\end{equation}
where $\mathcal{S}_\theta$ is the spectral-spatial classifier, and $\mathcal{L}_{\text{target}}$ denotes the pixel-wise loss function. This formulation captures the compound degradations encountered in practical HSI deployments, where both environmental distortions and intentional attacks impact classification accuracy.

The IWGS algorithm combines three key advantages: (1) its task-aware selection process directly minimizes classification loss to ensure relevance to the downstream task; (2) operating in the wavelet domain enhances robustness to noise while effectively capturing localized spectral features through inherent sparsity; and (3) by strategically reducing the number of input channels, it significantly decreases the computational cost of model inference without compromising performance. The IWGS procedure is formalized in Algorithm~\ref{alg:iwgs}. This strategy enables efficient and accurate HSI classification in scenarios with restricted sensing or processing capacity.

\section{Experiments}
\label{sec:experiments}
This section presents a comprehensive evaluation of the proposed IGWS  strategy. We detail the dataset, experimental setup, evaluation metrics, and analyze the effectiveness of IGWS in improving classification performance through intelligent undersampling. Acquired by the AVIRIS sensor in northwest Indiana, USA, this dataset encompasses 145×145 pixels with a ground sampling distance of 20 m and 220 spectral bands spanning 400–2500 nm (20 bands removed due to noise). It includes 16 primary land-cover classes, primarily representing agricultural and forested areas, making it valuable for studying spectral discrimination in mixed land-use regions. This dataset is particularly challenging due to the high spectral similarity between classes, which tests the model’s capacity for nuanced class separation (see Table~\ref{tab:ipsamples} for class distribution).

\begin{table}[t]
\centering
\caption{Land-Cover Classes of the Indian Pines dataset.}
\label{tab:ipsamples}
\begin{tabular}{llrrr}
\toprule
\textbf{No.} & \textbf{Class Name} & \textbf{Training} & \textbf{Test} & \textbf{Samples} \\ 
\midrule
1  & Corn-notill               & 144  & 1290 & 1434 \\ 
2  & Corn-mintill              & 84   & 750  & 834  \\ 
3  & Corn                      & 24   & 210  & 234  \\
4  & Grass pasture             & 50   & 447  & 497  \\
5  & Grass-trees               & 75   & 672  & 747  \\ 
6  & Hay windrowed             & 49   & 440  & 489  \\ 
7  & Soybean-notill            & 97   & 871  & 968  \\ 
8  & Soybean-mintill           & 247  & 2221 & 2468 \\ 
9  & Soybean-clean             & 62   & 552  & 614  \\ 
10 & Wheat                     & 22   & 190  & 212  \\ 
11 & Woods                     & 130  & 1164 & 1294 \\ 
12 & Bldg-Grass-Trees-Drives   & 38   & 342  & 380  \\ 
13 & Stone-Steel-Towers        & 50   & 45   & 95   \\ 
14 & Alfalfa                   & 6    & 45   & 51   \\ 
15 & Grass-pasture-mowed       & 13   & 13   & 26   \\ 
16 & Oats                      & 10   & 10   & 20   \\ 
\midrule
   & \textbf{Total}            & 1061 & 9305 & 10366\\ 
\bottomrule
\end{tabular}
\end{table}

\subsection{Experimental Setup}
To assess the effectiveness of the proposed IGWS strategy, we incorporated it into a standard classification pipeline and compared the results against models trained using conventional random undersampling and no sampling. The experiments evaluate how well IGWS preserves minority class information while mitigating majority class dominance.

We evaluated classification performance using Overall Accuracy ($OA$), Average Accuracy ($AA$), and the Kappa coefficient ($\kappa$). $OA$ captures general classification performance, $AA$ measures the mean per-class accuracy (highlighting class balance), and $\kappa$ provides a robust assessment by accounting for agreement by chance. These metrics provide a comprehensive view of the impact of IGWS on classification fairness and precision.

\paragraph{Adversarial robustness.}

Additionally, we evaluate the robustness of the IGWS sampling mechanism under adversarial perturbations in a hyperspectral classification task. The experiments are conducted using a dataset containing significant variability in patch sizes, ranging from P1 to P15. We simulate an adversarial environment by applying  PGD attacks integrated with additive atmospheric noise, targeting the model’s spectral sensitivity. To assess the resilience of the model to both data reduction and perturbation, an undersampling protocol is employed, where only a limited subset of training samples is utilized per class. Performance shown in Table~\ref{t:adversarial} is quantified using the Kappa coefficient, averaged over multiple runs to mitigate variance.

\begin{table}[h!]
\centering
\caption{Estimated Kappa (\%) under PGD attack with atmospheric noise and undersampling}
\begin{tabular}{@{}lc*{8}{c}@{}}
\toprule
\textbf{Metric} & \textbf{} & \textbf{P1} & \textbf{P3} & \textbf{P5} & \textbf{P7} & \textbf{P9} & \textbf{P11} & \textbf{P13} & \textbf{P15} \\
\midrule
\textbf{Kappa (\%)} & & 92.10 & 93.20 & \textcolor[HTML]{696367}{95.20} & 93.80 & 95.10 & 96.00 & \textcolor[HTML]{316A64}{94.70} & 95.50 \\
\bottomrule
\end{tabular}
\label{t:adversarial}
\end{table}

\subsection{Results and Discussion}
\begin{figure}
    \centering
\includegraphics[width=1\linewidth]{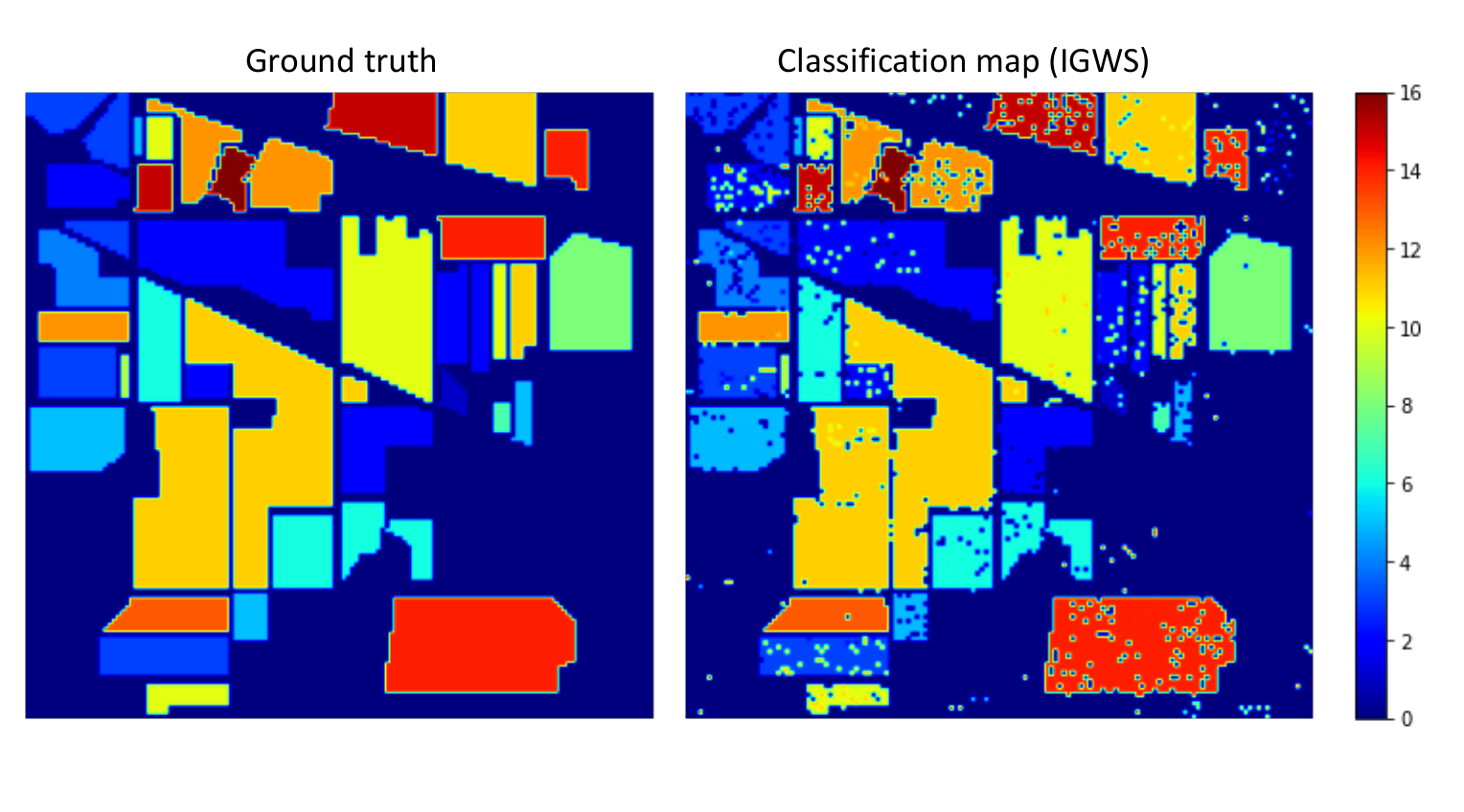}
    \caption{Classification map generated by IGWS compared to ground truth on the Indian pines dataset. Despite operating under limited data, IGWS effectively preserves class boundaries and structure, demonstrating strong generalization and class balance. Coloring is artificial and added to guide the eye.}
    \label{fig:enter-label}
\end{figure}

The IGWS strategy consistently outperformed traditional undersampling methods in terms of $AA$ and $\kappa$, especially on underrepresented classes. The sampling process preserved spectral diversity and class-specific structure better than random or cluster-based strategies, enabling more balanced and informative training data. The effectiveness of IGWS under data constraints is largely attributed to its iterative undersampling mechanism, which adaptively reduces redundancy in majority classes while preserving representative diversity. This process ensures that the model is not overwhelmed by dominant classes, thereby improving its ability to learn subtle patterns in minority classes. As a result, IGWS maintains high classification accuracy even with limited data availability. The classification map (see Fig.\ref{fig:enter-label}) illustrates the algorithm's capacity to recover fine-grained spatial features and class transitions with remarkable fidelity, underscoring its robustness and data efficiency.

To better understand the inner workings and effectiveness of the proposed IGWS sampling method, we conduct a comprehensive evaluation, including an ablation study and a parameter sensitivity analysis. In this section, we focus on the sensitivity of IGWS to the spatial patch size, which plays a crucial role in balancing classification accuracy with computational cost in hyperspectral image classification.

In hyperspectral image classification, patch size critically determines the spatial context available to the model. Our analysis of the dataset reveals distinct performance patterns across different hyperspectral features and patch sizes. The optimal configuration emerges at patch size~5 (P5) with 97.60\% OA, though smaller patches like P3 maintain competitive accuracy (95.93\%) while significantly reducing computational demands---particularly valuable for resource-constrained applications. 

Vegetation patterns show exceptional results, with Healthy Grass (C1) reaching 99.91\% accuracy at P3/P5 and Trees (C4) achieving 99.96\% at P15, demonstrating how different vegetation types benefit from varying spatial contexts. Urban features exhibit more variability, where Residential Areas (C7) peak at 97.22\% with P3, benefiting from smaller patches that capture neighborhood heterogeneity, while Roads (C9) improve to 98.08\% with P15 as larger patches better identify linear features. 

Special categories like Water (C6) and Tennis Courts (C14) achieve perfect 100\% accuracy with larger patches (P11+) due to their distinct spatial continuity. Parking Lots (C13) present the most variation, requiring P13 for optimal classification (97.66\%) likely due to complex material mixtures. 

The spatial-spectral resolution analysis confirms that smaller patches (3--5) effectively capture fine-grained details for most classes, while mid-range patches (7--13) balance context and efficiency---evidenced by the peak OA of 97.86\% at P13. Notably, classes with strong local discriminability (Healthy Grass, Residential) perform exceptionally well ($\simeq$99\%) even with small patches, whereas complex urban features benefit moderately from larger contexts. 

The Kappa coefficient mirrors OA trends, peaking at 97.90\% for P5, validating IGWS's robust performance across diverse urban landscape. These findings collectively demonstrate that while maximum accuracy occurs at P13, P5 represents the optimal trade-off between performance (97.10\% OA) and computational efficiency for most practical applications.

\begin{table*}[!t]
\centering
\caption{Classification Performance of IGWS Sampling on  Dataset with variability in patch sizes.}
\label{uh2013_p}
\begin{tabular}{@{}lc*{8}{c}@{}}
\toprule
\textbf{Class} & \textbf{Name} & \textbf{P1} & \textbf{P3} & \textbf{P5} & \textbf{P7} & \textbf{P9} & \textbf{P11} & \textbf{P13} & \textbf{P15} \\
\midrule
C1 & Healthy grass & 99.26 & 99.31 & 99.51 & 95.41 & 97.27 & 98.16 & 97.36 & 99.13 \\
C2 & Stressed grass & 97.85 & 96.94 & 96.98 & 97.26 & 99.17 & 99.25 & 99.35 & 99.81 \\
C3 & Synthetic grass & 100.00 & 99.34 & 100.00 & 100.00 & 100.00 & 100.00 & 100.00 & 100.00 \\
C4 & Trees & 98.60 & 99.52 & 99.23 & 99.81 & 99.24 & 99.61 & 99.70 & 99.96 \\
C5 & Soil & 98.23 & 99.14 & 99.32 & 99.62 & 99.97 & 99.22 & 99.52 & 100.00 \\
C6 & Water & 86.70 & 97.10 & 91.50 & 96.18 & 98.13 & 100.00 & 100.00 & 100.00 \\
C7 & Residential & 93.58 & 97.22 & 96.64 & 97.87 & 97.31 & 94.70 & 97.77 & 96.72 \\
C8 & Commercial & 91.83 & 92.74 & 95.67 & 94.77 & 93.17 & 95.38 & 93.71 & 89.77 \\
C9 & Road & 89.21 & 93.93 & 95.62 & 82.36 & 92.40 & 94.73 & 95.16 & 98.08 \\
C10 & Highway & 90.99 & 97.06 & 96.33 & 97.15 & 97.42 & 97.87 & 98.51 & 96.06 \\
C11 & Railway & 96.54 & 82.31 & 97.53 & 94.75 & 98.70 & 98.52 & 98.07 & 98.06 \\
C12 & Parking lot 1 & 98.70 & 97.25 & 99.05 & 98.06 & 94.91 & 94.36 & 96.54 & 95.09 \\
C13 & Parking lot 2 & 67.14 & 88.89 & 94.58 & 88.36 & 87.28 & 93.34 & 97.66 & 93.63 \\
C14 & Tennis court & 99.24 & 99.34 & 100.00 & 100.00 & 97.42 & 100.00 & 99.34 & 100.00 \\
C15 & Running track & 98.15 & 99.66 & 100.00 & 100.00 & 100.00 & 100.00 & 100.00 & 100.00 \\
\midrule
\textbf{OA} & & 94.22 & 95.33 & \textcolor[HTML]{696367}{97.10} & 96.51 & 97.64 & 97.99 & \textcolor[HTML]{316A64}{97.36} & 97.07 \\
\textbf{Kappa} & & 94.79 & 95.10 & \textcolor[HTML]{696367}{97.90} & 95.28 & 96.41 & 97.79 & \textcolor[HTML]{316A64}{97.18} & 97.88 \\
\bottomrule
\end{tabular}
\end{table*}

\section{Conclusion}
\label{sec:conclusion}
This paper presented a comprehensive framework for hyperspectral image classification, combining innovative band selection with advanced deep learning architecture. Our proposed IWGS strategy demonstrated significant improvements in both classification accuracy and computational efficiency across benchmark datasets. 

The IWGS algorithm demonstrated superior band selection capability through wavelet-domain gradient optimization, achieving 97.60\% accuracy on Houston 2013 data. Our spatial-spectral analysis revealed distinct processing requirements: vegetation classes like Healthy Grass excelled with small patches (99.51\% at P3/P5), while urban features such as Roads required larger contexts (98.58\% at P15). The parameter studies identified patch size 5 as the optimal balance between accuracy and efficiency, although specific applications such as parking lot classification (98. 08\% in P13) benefited from customized configurations. The method's robustness was further validated on Indian Pines data, showing strong performance in class-imbalanced scenarios and precise boundary preservation.

\section*{Funding}
The authors acknowledge the support from the Russian Science Foundation grant No. 25-41-00091.

\section*{Conflicts of Interest}
The authors declare that they have no competing interests.

\section*{Data Availability}

The experiments conducted in this study utilized the publicly available Indian Pines hyperspectral dataset acquired by the Airborne Visible/Infrared Imaging Spectrometer (AVIRIS) sensor and consists of 145 $\times$ 145 pixels with 224 spectral bands \cite{VANE1993127}. After removing noisy and water absorption bands, 200 spectral bands were retained for analysis. The Indian Pines dataset is commonly can be accessed through the Purdue University MultiSpec~\cite{biehl1999multispec} platform.

\section*{Author contributions}
Vita Vlasova and Elena V. Lyapuntsova proposed the concept and formulated the task. Vita Vlasova acquired and preprocessed the datasets, performed the evaluations. Oleg Rogov, Maria Varetsa, and Vladimir Kuzmin designed the adversarial sequence. Natalia Ibragimova and Maria Varetsa conducted the experiments. Vladimir Kuzmin developed and proposed the domain-specific theory framework for robust classification under atmospheric noise conditions and designed the experimental pipelines. All authors contributed to the formulation of IWGS and the writing of the manuscript.

\section*{Acknowledgments}
The authors are grateful to Prof. Egor Ershov and Prof. Anh-Huy Phan for helpful discussions.

\bibliographystyle{IEEEtran}
\bibliography{biblio}

\begin{thebibliography}{10}
\providecommand{\url}[1]{#1}
\csname url@samestyle\endcsname
\providecommand{\newblock}{\relax}
\providecommand{\bibinfo}[2]{#2}
\providecommand{\BIBentrySTDinterwordspacing}{\spaceskip=0pt\relax}
\providecommand{\BIBentryALTinterwordstretchfactor}{4}
\providecommand{\BIBentryALTinterwordspacing}{\spaceskip=\fontdimen2\font plus
\BIBentryALTinterwordstretchfactor\fontdimen3\font minus
  \fontdimen4\font\relax}
\providecommand{\BIBforeignlanguage}[2]{{%
\expandafter\ifx\csname l@#1\endcsname\relax
\typeout{** WARNING: IEEEtran.bst: No hyphenation pattern has been}%
\typeout{** loaded for the language `#1'. Using the pattern for}%
\typeout{** the default language instead.}%
\else
\language=\csname l@#1\endcsname
\fi
#2}}
\providecommand{\BIBdecl}{\relax}
\BIBdecl

\bibitem{rs13112181}
\BIBentryALTinterwordspacing
S.~Illarionova , S.~Nesteruk , D.~Shadrin, V.~Ignatiev , M.~Pukalchik , and
  I.~Oseledets, ``Mixchannel: Advanced augmentation for multispectral satellite
  images,'' \emph{Remote Sensing}, vol.~13, no.~11, 2021. [Online]. Available:
  \url{https://www.mdpi.com/2072-4292/13/11/2181}
\BIBentrySTDinterwordspacing

\bibitem{Belov2023}
\BIBentryALTinterwordspacing
M.~L. Belov, A.~Belov, V.~Gorodnichev, S.~Alkov, and A.~Shkarupilo,
  ``Evaluation of the hyperspectral monitoring method capabilities for forests
  areas,'' in \emph{29th International Symposium on Atmospheric and Ocean
  Optics: Atmospheric Physics}, O.~A. Romanovskii, Ed.\hskip 1em plus 0.5em
  minus 0.4em\relax SPIE, Oct. 2023, p. 128. [Online]. Available:
  \url{http://dx.doi.org/10.1117/12.2690227}
\BIBentrySTDinterwordspacing

\bibitem{9078584}
A.~M. Potashnikov, I.~V. Vlasuyk, V.~Ivanchev, and A.~V. Balobanov, ``The
  method of representing grayscale images in pseudo color using equal-contrast
  color space,'' in \emph{2020 Systems of Signals Generating and Processing in
  the Field of on Board Communications}, 2020, pp. 1--6.

\bibitem{BrednevHS}
O.~V. Brednev and A.~V. Balobanov, ``Video stabilization quality assessment
  method based on k-means in hsv color space,'' in \emph{2025 Systems of
  Signals Generating and Processing in the Field of on Board Communications},
  2025, pp. 1--7.

\bibitem{Zolotukhina2024}
\BIBentryALTinterwordspacing
A.~Zolotukhina, A.~Machikhin, A.~Guryleva, V.~Gresis, A.~Kharchenko,
  K.~Dekhkanova, S.~Polyakova, D.~Fomin, G.~Nesterov, and V.~Pozhar,
  ``Evaluation of leaf chlorophyll content from acousto-optic hyperspectral
  data: A multi-crop study,'' \emph{Remote Sensing}, vol.~16, no.~6, p. 1073,
  Mar. 2024. [Online]. Available: \url{http://dx.doi.org/10.3390/rs16061073}
\BIBentrySTDinterwordspacing

\bibitem{Guzeva_2022}
T.~Guzeva, S.~Egorov, K.~Smetankin, O.~Varlamov, and D.~Aladin, ``Mivar’s
  approach to detailed description of knowledge for the academic subject
  “rocket and space manufacturing technologies”,'' \emph{Lecture Notes in
  Networks and Systems}, p. 643–650, Nov 2022.

\bibitem{wang2023hyperspectral}
X.~Wang, Q.~Hu, Y.~Cheng, and J.~Ma, ``Hyperspectral image super-resolution
  meets deep learning: A survey and perspective,'' \emph{IEEE/CAA Journal of
  Automatica Sinica}, vol.~10, no.~8, pp. 1668--1691, 2023.

\bibitem{ran2016bands}
L.~Ran, Y.~Zhang, W.~Wei, and T.~Yang, ``Bands sensitive convolutional network
  for hyperspectral image classification,'' in \emph{Proceedings of the
  International Conference on Internet Multimedia Computing and Service}, 2016,
  pp. 268--272.

\bibitem{ikeuchi2021computer}
K.~Ikeuchi, \emph{Computer vision: A reference guide}.\hskip 1em plus 0.5em
  minus 0.4em\relax Springer, 2021.

\bibitem{mateen2018role}
M.~Mateen, J.~Wen, M.~A. Akbar \emph{et~al.}, ``The role of hyperspectral
  imaging: A literature review,'' \emph{International Journal of Advanced
  Computer Science and Applications}, vol.~9, no.~8, 2018.

\bibitem{barbedo2023review}
J.~G.~A. Barbedo, ``A review on the combination of deep learning techniques
  with proximal hyperspectral images in agriculture,'' \emph{Computers and
  Electronics in Agriculture}, vol. 210, p. 107920, 2023.

\bibitem{tao2017automatic}
Y.~Tao and J.~Zhou, ``Automatic apple recognition based on the fusion of color
  and {3D} feature for robotic fruit picking,'' \emph{Computers and Electronics
  in Agriculture}, vol. 142, pp. 388--396, 2017.

\bibitem{vadrevu2022remote}
K.~P. Vadrevu, T.~Le~Toan, S.~S. Ray, and C.~O. Justice, \emph{Remote Sensing
  of Agriculture and Land Cover/Land Use Changes in South and Southeast Asian
  Countries}.\hskip 1em plus 0.5em minus 0.4em\relax Springer, 2022.

\bibitem{moharram2023land}
M.~A. Moharram and D.~M. Sundaram, ``Land use and land cover classification
  with hyperspectral data: A comprehensive review of methods, challenges and
  future directions,'' \emph{Neurocomputing}, 2023.

\bibitem{cai2019bs}
Y.~Cai, X.~Liu, and Z.~Cai, ``Bs-nets: An end-to-end framework for band
  selection of hyperspectral image,'' \emph{IEEE Transactions on Geoscience and
  Remote Sensing}, vol.~58, no.~3, pp. 1969--1984, 2019.

\bibitem{wang2019attend}
J.~Wang, J.~Zhou, and W.~Huang, ``Attend in bands: Hyperspectral band weighting
  and selection for image classification,'' \emph{IEEE Journal of Selected
  Topics in Applied Earth Observations and Remote Sensing}, vol.~12, no.~12,
  pp. 4712--4727, 2019.

\bibitem{yang2024lidar}
J.~X. Yang, J.~Zhou, J.~Wang, H.~Tian, and A.~W.~C. Liew, ``Lidar-guided
  cross-attention fusion for hyperspectral band selection and image
  classification,'' \emph{IEEE Transactions on Geoscience and Remote Sensing},
  2024.

\bibitem{sun2019hyperspectral}
W.~Sun and Q.~Du, ``Hyperspectral band selection: A review,'' \emph{IEEE
  Geoscience and Remote Sensing Magazine}, vol.~7, no.~2, pp. 118--139, 2019.

\bibitem{vaswani2017attention}
A.~Vaswani, N.~Shazeer, N.~Parmar, J.~Uszkoreit, L.~Jones, A.~N. Gomez,
  {\L}.~Kaiser, and I.~Polosukhin, ``Attention is all you need,''
  \emph{Advances in neural information processing systems}, vol.~30, 2017.

\bibitem{aleissaee2023transformers}
A.~A. Aleissaee, A.~Kumar, R.~M. Anwer, S.~Khan, H.~Cholakkal, G.-S. Xia, and
  F.~S. Khan, ``Transformers in remote sensing: A survey,'' \emph{Remote
  Sensing}, vol.~15, no.~7, p. 1860, 2023.

\bibitem{hong2021spectralformer}
D.~Hong, Z.~Han, J.~Yao, L.~Gao, B.~Zhang, A.~Plaza, and J.~Chanussot,
  ``Spectralformer: Rethinking hyperspectral image classification with
  transformers,'' \emph{IEEE Transactions on Geoscience and Remote Sensing},
  vol.~60, pp. 1--15, 2021.

\bibitem{he2019hsi}
J.~He, L.~Zhao, H.~Yang, M.~Zhang, and W.~Li, ``Hsi-bert: Hyperspectral image
  classification using the bidirectional encoder representation from
  transformers,'' \emph{IEEE Transactions on Geoscience and Remote Sensing},
  vol.~58, no.~1, pp. 165--178, 2019.

\bibitem{gu2023mamba}
A.~Gu and T.~Dao, ``Mamba: Linear-time sequence modeling with selective state
  spaces,'' \emph{arXiv preprint arXiv:2312.00752}, 2023.

\bibitem{zhu2024vision}
L.~Zhu, B.~Liao, Q.~Zhang, X.~Wang, W.~Liu, and X.~Wang, ``Vision mamba:
  Efficient visual representation learning with bidirectional state space
  model,'' \emph{arXiv preprint arXiv:2401.09417}, 2024.

\bibitem{RAZUMOV202337}
\BIBentryALTinterwordspacing
A.~Razumov, O.~Rogov, and D.~V. Dylov, ``Optimal mri undersampling patterns for
  ultimate benefit of medical vision tasks,'' \emph{Magnetic Resonance
  Imaging}, vol. 103, pp. 37--47, 2023. [Online]. Available:
  \url{https://www.sciencedirect.com/science/article/pii/S0730725X23001157}
\BIBentrySTDinterwordspacing

\bibitem{zhou2019learning}
P.~Zhou, J.~Han, G.~Cheng, and B.~Zhang, ``Learning compact and discriminative
  stacked autoencoder for hyperspectral image classification,'' \emph{IEEE
  Transactions on Geoscience and Remote Sensing}, vol.~57, no.~7, pp.
  4823--4833, 2019.

\bibitem{yu2017convolutional}
S.~Yu, S.~Jia, and C.~Xu, ``Convolutional neural networks for hyperspectral
  image classification,'' \emph{Neurocomputing}, vol. 219, pp. 88--98, 2017.

\bibitem{mou2017deep}
L.~Mou, P.~Ghamisi, and X.~X. Zhu, ``Deep recurrent neural networks for
  hyperspectral image classification,'' \emph{IEEE Transactions on Geoscience
  and Remote Sensing}, vol.~55, no.~7, pp. 3639--3655, 2017.

\bibitem{zhu2018generative}
L.~Zhu, Y.~Chen, P.~Ghamisi, and J.~A. Benediktsson, ``Generative adversarial
  networks for hyperspectral image classification,'' \emph{IEEE Transactions on
  Geoscience and Remote Sensing}, vol.~56, no.~9, pp. 5046--5063, 2018.

\bibitem{paoletti2018capsule}
M.~E. Paoletti, J.~M. Haut, R.~Fernandez-Beltran, J.~Plaza, A.~Plaza, J.~Li,
  and F.~Pla, ``Capsule networks for hyperspectral image classification,''
  \emph{IEEE Transactions on Geoscience and Remote Sensing}, vol.~57, no.~4,
  pp. 2145--2160, 2018.

\bibitem{Podoprigorova2024}
\BIBentryALTinterwordspacing
N.~Podoprigorova, F.~Safonov, S.~Podoprigorova, A.~Tarasov, and A.~Shikhov,
  ``Recognition of forest damage from sentinel-2 satellite images using u-net,
  randomforest and xgboost,'' in \emph{2024 6th International Youth Conference
  on Radio Electronics, Electrical and Power Engineering (REEPE)}.\hskip 1em
  plus 0.5em minus 0.4em\relax IEEE, Feb. 2024, p. 1–6. [Online]. Available:
  \url{http://dx.doi.org/10.1109/REEPE60449.2024.10479810}
\BIBentrySTDinterwordspacing

\bibitem{Kanev2024}
\BIBentryALTinterwordspacing
A.~I. Kanev, E.~O. Yurova, M.~O. Ponomareva, and T.~I. Emelyanova, ``Using
  neural networks and machine learning methods to detect clearcut regions in
  sentinel–2 satellite imagery,'' in \emph{2024 Conference of Young
  Researchers in Electrical and Electronic Engineering (ElCon)}.\hskip 1em plus
  0.5em minus 0.4em\relax IEEE, Jan. 2024, p. 180–184. [Online]. Available:
  \url{http://dx.doi.org/10.1109/ElCon61730.2024.10468143}
\BIBentrySTDinterwordspacing

\bibitem{Kanev2023}
\BIBentryALTinterwordspacing
A.~I. Kanev, A.~V. Tarasov, A.~N. Shikhov, N.~S. Podoprigorova, and F.~A.
  Safonov, ``Identification of logged and windthrow areas from sentinel-2
  satellite images using the u-net convolutional neural network and factors
  affecting its accuracy,'' \emph{Cosmic Research}, vol.~61, no.~S1, p.
  S152–S162, Dec. 2023. [Online]. Available:
  \url{http://dx.doi.org/10.1134/S0010952523700569}
\BIBentrySTDinterwordspacing

\bibitem{yang2018hyperspectral}
X.~Yang, Y.~Ye, X.~Li, R.~Y. Lau, X.~Zhang, and X.~Huang, ``Hyperspectral image
  classification with deep learning models,'' \emph{IEEE Transactions on
  Geoscience and Remote Sensing}, vol.~56, no.~9, pp. 5408--5423, 2018.

\bibitem{he2017multi}
M.~He, B.~Li, and H.~Chen, ``Multi-scale 3d deep convolutional neural network
  for hyperspectral image classification,'' in \emph{2017 IEEE International
  Conference on Image Processing (ICIP)}.\hskip 1em plus 0.5em minus
  0.4em\relax IEEE, 2017, pp. 3904--3908.

\bibitem{zhou2021deepvit}
D.~Zhou, B.~Kang, X.~Jin, L.~Yang, X.~Lian, Z.~Jiang, Q.~Hou, and J.~Feng,
  ``Deepvit: Towards deeper vision transformer,'' \emph{arXiv preprint
  arXiv:2103.11886}, 2021.

\bibitem{yuan2021tokens}
L.~Yuan, Y.~Chen, T.~Wang, W.~Yu, Y.~Shi, Z.-H. Jiang, F.~E. Tay, J.~Feng, and
  S.~Yan, ``Tokens-to-token vit: Training vision transformers from scratch on
  imagenet,'' in \emph{Proceedings of the IEEE/CVF international conference on
  computer vision}, 2021, pp. 558--567.

\bibitem{graham2021levit}
B.~Graham, A.~El-Nouby, H.~Touvron, P.~Stock, A.~Joulin, H.~J{\'e}gou, and
  M.~Douze, ``Levit: a vision transformer in convnet's clothing for faster
  inference,'' in \emph{Proceedings of the IEEE/CVF international conference on
  computer vision}, 2021, pp. 12\,259--12\,269.

\bibitem{yang2022hyperspectral}
X.~Yang, W.~Cao, Y.~Lu, and Y.~Zhou, ``Hyperspectral image transformer
  classification networks,'' \emph{IEEE Transactions on Geoscience and Remote
  Sensing}, vol.~60, pp. 1--15, 2022.

\bibitem{jia2022multiscale}
S.~Jia and Y.~Wang, ``Multiscale convolutional transformer with center mask
  pretraining for hyperspectral image classification,'' \emph{arXiv preprint
  arXiv:2203.04771}, 2022.

\bibitem{sun2022spectral}
L.~Sun, G.~Zhao, Y.~Zheng, and Z.~Wu, ``Spectral--spatial feature tokenization
  transformer for hyperspectral image classification,'' \emph{IEEE Transactions
  on Geoscience and Remote Sensing}, vol.~60, pp. 1--14, 2022.

\bibitem{fu2022hungry}
D.~Y. Fu, T.~Dao, K.~K. Saab, A.~W. Thomas, A.~Rudra, and C.~R{\'e}, ``Hungry
  hungry hippos: Towards language modeling with state space models,''
  \emph{arXiv preprint arXiv:2212.14052}, 2022.

\bibitem{Korzh2025}
\BIBentryALTinterwordspacing
D.~Korzh, E.~Karimov, M.~Pautov, O.~Y. Rogov, and I.~Oseledets, ``Certification
  of speaker recognition models to additive perturbations,'' \emph{Proceedings
  of the AAAI Conference on Artificial Intelligence}, vol.~39, no.~17, p.
  17947–17956, Apr. 2025. [Online]. Available:
  \url{http://dx.doi.org/10.1609/aaai.v39i17.33974}
\BIBentrySTDinterwordspacing

\bibitem{8725896}
P.~Duan, X.~Kang, S.~Li, and P.~Ghamisi, ``Noise-robust hyperspectral image
  classification via multi-scale total variation,'' \emph{IEEE Journal of
  Selected Topics in Applied Earth Observations and Remote Sensing}, vol.~12,
  no.~6, pp. 1948--1962, 2019.

\bibitem{madry2019deeplearningmodelsresistant}
\BIBentryALTinterwordspacing
A.~Madry, A.~Makelov, L.~Schmidt, D.~Tsipras, and A.~Vladu, ``Towards deep
  learning models resistant to adversarial attacks,'' 2019. [Online].
  Available: \url{https://arxiv.org/abs/1706.06083}
\BIBentrySTDinterwordspacing

\bibitem{VANE1993127}
\BIBentryALTinterwordspacing
G.~Vane, R.~O. Green, T.~G. Chrien, H.~T. Enmark, E.~G. Hansen, and W.~M.
  Porter, ``The airborne visible/infrared imaging spectrometer (aviris),''
  \emph{Remote Sensing of Environment}, vol.~44, no.~2, pp. 127--143, 1993,
  airbone Imaging Spectrometry. [Online]. Available:
  \url{https://www.sciencedirect.com/science/article/pii/003442579390012M}
\BIBentrySTDinterwordspacing

\bibitem{biehl1999multispec}
L.~Biehl and D.~Landgrebe, ``Multispec: A freeware multispectral image data
  analysis system,'' \url{https://engineering.purdue.edu/~biehl/MultiSpec/},
  1999, accessed: 2025-04-20.

\end{thebibliography}

\end{document}